# Patterns of near-crash events in a naturalistic driving dataset: applying rules mining

**Xiaoqiang "Jack" Kong[a], *, Subasish Das, Ph.D.[b], Hongmin "Tracy" Zhou, Ph.D.[b], Yunlong Zhang, Ph.D.[a]**

[*] Corresponding Author, Email: x-kong@tamu.edu
[a] Texas A&M University, 3136 TAMU, College Station, TX 77843
[b] Texas A&M Transportation Institute, 1111 RELLIS Parkway, Bryan, TX  77807

August 2020

Revised for Possible Publication in Accident Analysis and Prevention

## ABSTRACT

This study aims to explore the associations between near-crash events and road geometry and trip features by investigating a naturalistic driving dataset and a corresponding roadway inventory dataset using an association rule mining method – the Apriori algorithm. To provide more insights into near-crash behavior, this study classified near-crash events into two severity levels: trivial near-crash events (-7.5 g ≤ deceleration rate ≤ -4.5 g) and non-trivial near-crash events (≤ -7.5 g). Each category for all variables is considered an item, and a set of items is considered an itemset. From the perspective of descriptive statistics, the frequency of the itemsets generated by the Apriori algorithm suggests that near-crash events are highly associated with several factors, including roadways without access control, driving during non-peak hours, roadways without a shoulder or a median, roadways with the minor arterial functional class, and roadways with a speed limit between 30 and 60 mph. By comparing the frequency of the occurrence of the itemset during trivial and non-trivial near-crash events, the results indicate that the length of the trip is a strong indicator of the near-crash event type. The results show that non-trivial near-crash events are more likely to occur if the trip is longer than 2 hours. After applying the association rule mining algorithm, more interesting patterns for the two near-crash events were generated through the rules. The main findings include: 1) trivial near-crash events are more likely to occur on roadways without a median and shoulder that have a relatively lower functional class; 2) relatively higher functional roadways with relatively wide medians and shoulders could be an intriguing combination for non-trivial near-crash events; 3) non-trivial near-crash events often occur on long trips (more than 2 hours); 4) congestion on roadways that have a lower functional class is a dominant rule of non-trivial near-crash events. This study associates near-crash events and the corresponding road geometry and trip features to provide a unique understanding of near-crash events.

## 1. Introduction

While numerous efforts over the years have been devoted to reducing crash fatalities, there is still a considerable cost associated with traffic crashes. The estimated economic cost of all fatalities due to traffic crashes in 2018 was approximately $55 billion in the United States (CDC, 2020). Such a huge cost warrants continued investigation into the contributing factors of crash fatalities and the implementation of effective countermeasures for improving traffic safety. Traditional safety studies have generally focused on identifying correlations between crashes and roadway features. Due to a lack of substantial driving behavior information in conventional historical crash datasets, these studies can seldom identify driving behaviors that contribute to crashes. Moreover, traditional studies require crash data spanning an extended period of time. Such data would not be available for newly constructed roadways.

With the recent advancements in data collection technologies, naturalistic driving studies (NDS) have gained attention and provided a valuable surrogate resource, near-crash events, to the safety analysis. These studies collect trajectory data of a number of vehicles over an extended period in an attempt to capture driving behaviors, along with the surrounding environment information, immediately prior to crash or near-crash incidents (Hankey et al., 2016). Moreover, with the advent of the connected vehicle (CV) technology, vehicle trajectory data is expected to become more accessible. Various surrogate safety measures have been proposed for identifying risky driving behaviors using vehicle trajectory data. Safety assessment using trajectory data and surrogate safety measures provides a good alternative for identifying contributing factors leading to a crash under such conditions. Criteria derived from associating these surrogate safety measures with crashes can help detect hazardous events in real-time, which can be used in various crash-avoidance applications to improve traffic safety.

This study aims to identify the association between contribution factors and detect near-crashes using a naturalistic driving dataset. To ensure sufficient data availability, the authors focused on roadway geometry and trip features for potential attributable factors and longitudinal near-crash event. Using thresholds of surrogate safety measures found in the literature, this study differentiates near-crash events into trivial and non-trivial near-crash events. The association rule algorithm Apriori is then used to mine the association rules for the two types of events using data NDS data and road network inventory data. Mining rules for near-crash events with different severity labels can not only provide insights into the associations between the near-crash events and geometric features/trip features, but also offer additional insights by comparing the association rules for the two types of near-crash events. The following section provides a review of the literature on surrogate safety measures to help identify near-crash events.

## 2. Literature Review

A majority of the surrogate safety measures are derived from vehicle kinematic characteristics, which can be categorized into time-based measures, speed-based measures, and acceleration-based measures. The time to collision (TTC) is a widely used time-based measure (González et al., 2014) and is originated from the traffic conflict technique (Glauz and Migletz, 1980; Parker and Zegeer, 1989; Zegeer, 1986). The TTC is the time to collide if two consecutive vehicles continue at their present speed and along the same path (Gettman and Head, 2003). A threshold of 4 sec for TTC has been suggested to identify safe and uncomfortable situations (Hydén and Linderholm, 1984), and a TTC value of 1.5 sec or lower has been associated with more severe conflicts (Minderhoud and Bovy, 2001). Because TTC does not consider the change in speed due to the driver's reaction to the potential collision, the modified TTC (MTTC) measures that indirectly consider the relative speed and acceleration between vehicles have been used when TTC is lower than the specified threshold values (He et al., 2018; Minderhoud and Bovy, 2001; Ozbay et al., 2008). The percent of time tailgating (headway shorter than a predefined threshold, e.g., 0.4 sec) is another time-based measure used to identify aggressive drivers who tend to be involved in more near-crash or crash incidents (Feng et al., 2017). Conventional safety analysis on crash data has found that speed not only affects the severity of a crash, but is also related to the risk of a vehicle being involved in a crash (Elvik et al., 2004). Therefore, various speed related measures have been derived from vehicle trajectory data as a safety indicator. The maximum speed and the maximum speed difference between two consecutive vehicles involved in a conflict event are two speed based measures used to identify the severity of a collision (Gettman and Head, 2003). The percent of the time a vehicle traveling above a predefined threshold, e.g., 85 mph (Hydén and Linderholm, 1984), and the speed distribution of a vehicle (González et al., 2014) have been used to identify aggressive drivers. Vehicular speed profiles generated from simulator trajectory data are found to be a good indicator of driving risks when combined with roadway geometry data (Wang et al., 2010).

The frequency of braking with a negative jerk, indicating a greater change rate of longitudinal deceleration, is found to be an effective measure of aggressive drivers (Ozbay et al., 2008). The standard deviations of longitudinal and lateral acceleration have been used as a measure to detect abnormal driving instances such as weaving, swerving, side slipping, fast U-turn, turning with a wide radius, and sudden braking (Chen et al., 2015). Thresholds of hard braking (longitudinal deceleration rate less than -0.45g), rapid starts (longitudinal acceleration rate greater than 0.35g), and swerving (absolute lateral turn rate greater than 0.50g) have been identified to detect these elevated G-force events as the kinematic risky driving behaviors (Simons-Morton et al., 2013). Acceleration thresholds changing with speeds to account for different driving situations have also been used to identify extreme events (Liu and Khattak, 2016).

Various thresholds of the longitudinal deceleration rate have been used in the literature for identifying near-crash events. A study on risky driving behaviors among younger and older drivers using the Strategic Highway Research Program 2 (SHRP2) naturalistic driving study (NDS) data supported the use of a threshold value at -0.45g for classifying risky driving behaviors (Simons-Morton et al., 2019). The threshold of -0.45 g is also used for identifying extreme events in the American Association of State Highway and Transportation Officials (AASHTO) Greenbook (Hancock and Wright, 2013). The longitudinal deceleration with a value of -0.6g has been used as the threshold to identify potential crashes and near-crashes (Lin et al., 2008; Ozbay et al., 2008; Perez et al., 2017). Research on the 100-Car NDS data established a set of criteria for detecting candidate crashes, one of which is longitudinal deceleration less or equal to -0.6g (Dingus et al., 2006). A lightly higher deceleration rate of -0.65g was used for assessing teen driving safety (Lee et al., 2011) and for initial screening of the SHRP2 NDS data for crash and near-crash events (Hankey et al., 2016). Yet another study on the kinematic threshold of crash and near-crash by using data from the SHRP 2 NDS and the Canada Naturalistic Driving Study (CNDS) suggested an even higher deceleration threshold of -0.75g could improve the data screening performance (Perez et al., 2017). A more recent study compared deceleration rates between crash and near-crash events using the SHRP 2 NDS data and found that drivers in near-crash events on average decelerated 0.018g quicker than drivers in equivalent crashes (Wood and Zhang, 2021)

As discussed above, the existing studies all agreed that near-crash events could be identified through the rate of deceleration. However, studies across the time have adopted different thresholds of longitudinal deceleration rates ranging from -0.75g to -0.45g to identify near-crash events. Therefore, in this study, events with a deceleration rate less than -0.45g are identified as near-crash events, which is considered a weak criterion. The value of -0.75g is used to further differentiate two types of near-crash events: events with a deceleration rate between -0.45g and -0.75g are labeled as trivial near-crash events, whereas events with a deceleration rate less than -0.75g are labeled as non-trivial near-crash events. The literature review indicates that there is a need for an in-depth investigation of the patterns of trivial and non-trivial near-crash events. The current will address the current limitation by applying association rules mining on a unique dataset.

## 3. Data Preparation

Two datasets have been utilized for the analysis. One is the naturalistic driving data called Safety Pilot Model Deployment (SPMD) data, which was collected through the "Connected Vehicle Safety Pilot Program" (Hamilton, 2015; USDOT, 2020). The SPMD dataset contains detailed naturalistic driving information including, but not limited to, trip trajectory, driving maneuver (e.g., longitudinal acceleration), and driving environment (including distance and movements of surrounding vehicles). Another dataset is the Michigan road network inventory data called the Highway Performance Monitoring System (HPMS). This dataset contains roadway geometry information, characteristics, and average annual daily traffic (AADT).

### 3.1. Safety Pilot Model Deployment (SPMD) Data

The SPMD collected vehicles' trajectory data and the driving environment data using a Data Acquisition System 1 (DAS1). The raw data are available on DATA.GOV website (USDOT, 2020). The data was collected in Michigan from October 2012 to April 2013. Table 1 shows the SPMD data variables used in this study. The unit of longitudinal acceleration is m/sec$^2$. To be consistent with the literature, this unit is converted to $g$ in our study.

**Table 1. SPMD Data Variable**

| Variable | Unit | Description | Dataset in SPMD |
|---|---|---|---|
| Device | -- | Unique identifier for a device/ vehicle. | DataFrontTargets and DataWsu |
| Trip | -- | A trip starts when ignition is ON and ends when it is OFF. Unique for a device/vehicle. | DataFrontTargets and DataWsu |
| Time | Centiseconds | Time in centiseconds since DAS started for a trip (generally start of ignition) | DataFrontTargets and DataWsu |
| CIPV | -- | 1, If the identified vehicle is the closest in a vehicle's path | DataFrontTargets |
| Range | m | Spacing between subject and target vehicle | DataFrontTargets |
| Transversal | m | Lateral distance between the subject and target vehicle | DataFrontTargets |
| Rangerate | m/s | Relative speed between the subject and target vehicle | DataFrontTargets |

| Status | -- | Status of the target vehicle (Moving, standing, stopped, oncoming, parked, and unknown) | DataFrontTargets |
|---|---|---|---|
| TargetType | -- | Type of target vehicle (Car, truck, motorcycle, pedestrian, and bicycle) | DataFrontTargets |
| LatitudeWsu | Degree | Latitude for WSU receiver | DataWsu |
| LongitudeWsu | Degree | Longitude for WSU receiver | DataWsu |
| GpsSpeedWsu | m/sec | Speed from WSU GPS receiver | DataWsu |
| AxWsu | m/sec$^2$ | Longitudinal acceleration from the CAN Bus via WSU | DataWsu |

*Note: AxWsu unit 'm/sec$^2$' is converted to 'g' during the data cleaning process*

To ensure the reliability of the results of this study, data records meeting the following criteria are excluded from the dataset:

- Trips with a duration shorter than 10 minutes, i.e., Time ≤ 60,000 centiseconds;
- Scenarios with no vehicle in front of the subject vehicle; i.e., CIPV = 0; and
- Scenarios with a longitudinal distance between the subject and lead vehicles greater than 50 ft (Dingus et al., 2006, p. 100).

For many trips shorter than 10 minutes, the data presented a low rate of near-crash events, and trip trajectories are not on the road network HPMS system, which will lead to incomplete data. For the second criteria, CIPV equals zero means there is no vehicle ahead. Thus, there is no chance for near-crash events. For the last criteria, it is to rule out these near-crash events meet the longitude deceleration threshold, but it is still away from the lead vehicle. After the data cleaning process, data from 92 vehicles are available for the study. Ninety-two vehicles drove about 100,000 minutes and about 35,000 miles in the data collection period.

## 3.2. Highway Performance Monitoring System (HPMS)

The HPMS covers the inventory of major roadways for all the states in the U.S. The authors extracted the 2016 HPMS dataset for Michigan for this study. The dataset includes roadway features (e.g., the number of lanes and type of functional classification) and traffic features (e.g., AADT). Table 2 lists the roadway features included in this study. Some features, such as shoulder types, are not included in this study due to a lot of missing values.

**Table 2 HPMS Data Variables.**

| Variable | Description |
|---|---|
| f_system | Function system |
| access_con | Access control |
| shoulder_width | The width of the shoulder |
| aadt | Annual average traffic |
| median_type | The type of median |
| median_width | The width of the median |
| speed_limit | Speed limit (mph) |

3.3. Data Integration

The near-crash data points and HPMS shapefiles have no common identification key. Thus, the join two datasets together through a common key is not a viable option here. To integrate the two datasets, the spatial join function of QGIS was used. Near-crash events are datapoints with longitude and latitude coordinates. HPMS is the line shapefile with spatial coordinates. Therefore, two spatial files can be spatially joined by the nearest feature. For example, a near-crash event will capture its nearest HPMS road segment and merge the information of that road segment back into the near-crash database. The common error that might cause the incorrect integration is that the nearest road segment to some near-crash crash event might not the real segment where the near-crash occurred. For example, if a near-crash event occurred on a highway segment where has an overpass road. The overpass may be geographically closer to the overpass segment rather than the highway segment. The spatial join will definitely integrate the geometric information of that overpass segment back into the near-crash database. To decrease the possibility of the incorrect conflation, these road segments near the SPMD trajectories but not used by any trip in the PSMD dataset are manually deselected. Meanwhile, the trajectories with different vehicle IDs are conflated separately to avoid further possible wrong conflations.

The SPMD dataset has been processed before integrating with the HPMS inventory data. The longitudinal near-crash events are identified by the AxWsu value (Longitudinal acceleration rate) available in the SPMD dataset. As aforementioned, data points with an acceleration higher than – 4.45g were discarded. The rest of the data points are considered near-crash events and further classified as trivial near-crash events (-0.75g ≤ deceleration rate ≤ -0.45g) and non-trivial near-crash events (deceleration rate ≤ -0.75g). Moreover, the authors considered braking events that were at least three minutes apart as distinct events. This treatment can avoid duplicating near-crash events from the same incident when several hard break maneuvers might have occurred in a single near-crash situation. There are 957 near-crash events extracted from this naturalistic driving dataset. Table 3 presents the summary statistics of the cleaned near-crash dataset.

**Table 3 Count and Percentages of the Key Variables**

| Variables | Data Source | Levels | Count | Percentage |
|---|---|---|---|---|
| speed | SPMD | 30 - 60 mph | 235 | 24.56% |
| | | larger than 60 mph | 99 | 10.34% |
| | | less than 30 mph | 623 | 65.10% |
| functional class | HPMS | interstate | 64 | 6.69% |
| | | major collector | 61 | 6.37% |
| | | minor arterial | 575 | 60.08% |
| | | principal arterial | 257 | 26.85% |
| access_con | HPMS | no | 788 | 82.34% |
| | | yes | 169 | 17.66% |
| shoulder_width | HPMS | 4 - 8 ft | 95 | 9.93% |
| | | larger than 8 ft | 75 | 7.84% |
| | | less than 4 ft | 25 | 2.61% |
| | | no shoulder | 762 | 79.62% |
| lane_width | HPMS | 11 ft | 314 | 32.81% |
| | | 12 ft | 313 | 32.71% |

| | | 13 ft | 330 | 34.48% |
|---|---|---|---|---|
| median_type | HPMS | barrier | 73 | 7.63% |
| | | curbed | 347 | 36.26% |
| | | unprotected | 537 | 56.11% |
| median_width | HPMS | 35 - 60 ft | 386 | 40.33% |
| | | larger than 60 ft | 69 | 7.21% |
| | | less than 35 ft | 62 | 6.48% |
| | | no median | 440 | 45.98% |
| speed_limit | HPMS | less than 30 mph | 248 | 25.91% |
| | | 30 - 60 mph | 541 | 56.53% |
| | | larger than 60 mph | 168 | 17.55% |
| peak | SPMD | no | 700 | 73.15% |
| | | yes | 257 | 26.85% |
| aadt | HPMS | less than 20,000 vehicles per day (vpd) | 266 | 27.80% |
| | | 20,000 − 40,000 vpd | 250 | 54.34% |
| | | 40,000 − 70,000 vpd | 92 | 9.61% |
| | | more than 70,000 vpd | 79 | 8.25% |
| traveltime | SPMD | less than 20 minutes | 184 | 19.23% |
| | | 20 to 60 minutes | 288 | 30.09% |
| | | 1 - 2 hours | 132 | 13.79% |
| | | longer than 2 hours | 353 | 36.89% |
| nv_severity | SPMD | non-trivial | 401 | 41.90% |
| | | trivial | 556 | 58.10% |

## 4. Methodology

This study adopted the association rules mining method in identifying the relationship between near-crash events and geometry and trip features. This method aims to discover significant patterns when categorical data variables frequently occur together under certain conditions in large databases. Revealing these co-occurrence patterns in a database involves the use of machine learning models to identify the association rules. Three common performance measures are generally used in association rules mining. These performance metrics are *Support*, *Confidence*, and *Lift*. This method has been adopted in identifying association patterns for work zone crashes (Weng et al., 2016), hit-and-run crashes (Das et al., 2019), pedestrian crashes (Das et al., 2018), speeding behaviors (Kong et al., 2020), and crashes at urban roundabouts (Montella, 2011).

The Apriori algorithm (Agrawal and Srikant, 1994) was used in generating association rules in this study. The algorithm is a level-wise, breadth-first algorithm that counts transactions (e.g., crash level information in a row can be considered a transaction). It considers any subset of a frequent itemset (a set containing two or more items) to also be a frequent itemset. No superset of any infrequent itemset is generated or tested to allow pruning of many item combinations. This algorithm can be applied to several forms of itemsets, including frequent itemsets, maximal frequent itemsets, and closed frequent itemsets. It can help mine out frequently occurring itemsets, subsequences, arrangements, and interesting associations between various items. For the crash datasets of concern in this study, a set of definitions are provided here according to the Apriori algorithm:

Consider $I = \{i_1, i_2, \ldots i_m\}$ is a set of incident categories for a particular near-crash incidents (items). Let $C = \{c_1, c_2, \ldots, c_n\}$ be a set of near-crash records of incident information (transactions), where each near-crash record $c_i$ contains a subset of incident categories chosen from $I$. An itemset with $k$ items is called a $k$-itemset. An association rule can be demonstrated as $A \rightarrow B$, or Antecedent $\rightarrow$ Consequent, where $A$ and $B$ are disjoint incident categories. For example, in this study, the Antecedent could be one or a combination of items, such as { lane_width = 12 ft, median_type = unprotected, peak=no }, and the Consequent could be {near-crash severity = trivial}. The strength of an association rule can be measured using the values of the aforementioned performance measures. For the purpose of analysis in this study, *Support* is defined as the percentage of events in the dataset that contains certain itemset - incident categories. This itemset could be a single item or a set of items. The more items in the itemset, the occurrence of this itemset in the whole dataset is less frequent. Thus, the *support* value would become smaller. *Confidence* is the ratio of the number of near-crashes in $C$ to the number of near-crashes that involve all incident categories in $I$. In this case, the *confidence* can be understood as the occurrence of antecedent (a set of items) and consequent (a type of near-crash) together versus the occurrence of antecedent (a set of items) in the whole dataset. *Lift* measures the interdependence between the antecedent and consequent. A *Lift* value greater than one, i.e., $L$ $(A \rightarrow B) > 1$, indicates significant interdependence between the antecedent and the consequent, while a *Lift* value less than one, i.e., $L(A \rightarrow B) < 1$ indicates low interdependence. A *Lift* value of one, i.e., $L(A \rightarrow B) = 1$, indicatates independence. A rule with a single antecedent and a single consequent can be defined as a *2-product* or 2-*item* rule; similarly, a rule with two antecedents and a single consequent or one antecedent and two consequents is defined as a 3-product or 3-item rule. A critical inference of the association rules is that the generated rules need not be interpreted as association only (none of the rules will indicate any causation). The *Support* measure is determined using equation (1) through equation (3). *Confidence* and *Lift* is determined using equations (4) and equation (5), respectively.

$$S(A) = \frac{\sigma(A)}{N} \tag{1}$$

$$S(B) = \frac{\sigma(B)}{N} \tag{2}$$

$$S(A \rightarrow B) = \frac{\sigma(A \cap B)}{N} \tag{3}$$

$$C(A \rightarrow B) = \frac{S(A \rightarrow B)}{S(A)} \tag{4}$$

$$L(A \rightarrow B) = \frac{S(A \rightarrow B)}{S(A).S(B)} \tag{5}$$

where,
$\sigma(A)$ = Count of incidents with antecedent $A$
$\sigma(B)$ = Count of incidents with consequent $B$
$\sigma(A \cap B)$ = Count of incidents with both $A$ antecedent and $B$ consequent
$N$ = Total count of incidents
$S(A)$ = Support of antecedent
$S(B)$ = Support of consequent

$S (A \rightarrow B)$ = Support of the association rule $(A \rightarrow B)$
$C (A \rightarrow B)$ = Confidence of the association rule $(A \rightarrow B)$
$L (A \rightarrow B)$ = Lift of the association rule $(A \rightarrow B)$

After preparing the conflated dataset in a categorical format, the data is ready for analysis. A R package "arules" is utilized in this research. To perform association rules mining, the parameters of the algorithm need to be tuned in many iterations to obtain the best performance in order to mine meaningful association rules. The final parameters used in this study are minimum support equals 0.1, the minimum confidence equals 0.1, the min len is 3, and max len is 5. This parameter tuning process could suffer from subjectivity and needs extra attention to avoid potentially biased judgments. There are two essential criteria for this parameter tuning process to ensure the quality of association rules. The first one is the minimum support should be a reasonable value. The second one is that the rules reported in the analysis should be meaningful. The minimum support value ensures the pattern (an itemset) appearing in a relatively large amount of observations. For example, the minimum support value for the study (with 957 observations) is 0.1, which indicates, for any pattern (itemset), it must show in at least 0.1*957 = 95.7 observations. If a pattern is presented in about 100 out of 957 observations, this pattern should be significant and worth the reporting. If the minimum support value is set as 0.01, this small support value may generate more rules, but these rules (pattern) may only appear in 10 out of 957 observations. Therefore, with a pattern appearing in a relatively small amount of observations, that does not suggest an affirmative pattern. On the other hand, the many patterns may have high support value and frequently appears in the dataset, but they may not provide any new or meaningful information. For example, {functional class = interstate, median = unprotected} may frequently occur in the observations due to the nature of interstate highway. However, this pattern is obvious and not worth reporting.

## 5. Results and Findings

To mine association rules from the prepared dataset, it is important to convert the dataset into transaction format. Different categories of each feature are transformed as items.

### 5.1. Exploratory Analysis

Figure 1 shows the relative frequency of each item from the whole near-crash event dataset. There are 20 items in this figure with the highest frequency. As mentioned in the data description section, the SPMD data are from 100 participants over several months. It is reasonable to state that most of the geometric features in the original dataset should well represent their population distribution. Only events that match the longitudinal near-crash criteria are selected for this study. Thus, it is reasonable to believe that this frequency plot could actually reflect the real association between near-crash and roadway or trip features. Figure 1 indicates near-crash events are more likely to be associated with road segments without access control, absence of shoulders, driving during non-peak hours, minor arterials, trivial near-crash event, 30-60 mph posted speed limit, and unprotected median.

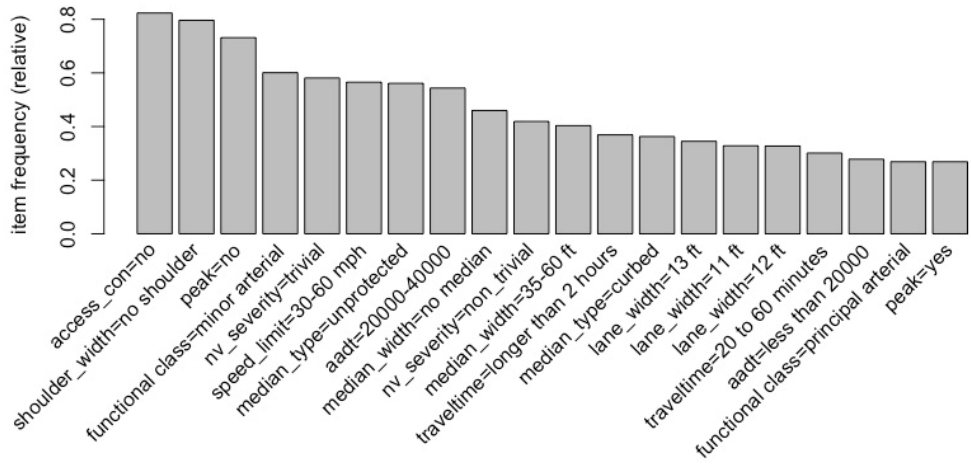

**Figure 1 Item Frequency Plot of the Whole Dataset**

Figure 2 and Figure 3 are item frequency plots for trivial near-crash events and non-trivial near-crash events, respectively. By comparing these two plots, there are similarities and differences in terms of frequent items. Both types of events are likely to be associated with roadways without access control, absence of shoulder, minor arterial roadways, and driving during non-peak hours. However, non-trivial near-crash events are more likely to be associated with trips longer than 2 hours and on the roadway with higher traffic volumes; trivial near-crash events are more likely to be associated with trips with a travel time of 20 – 60 minutes and on-road segments without a median.

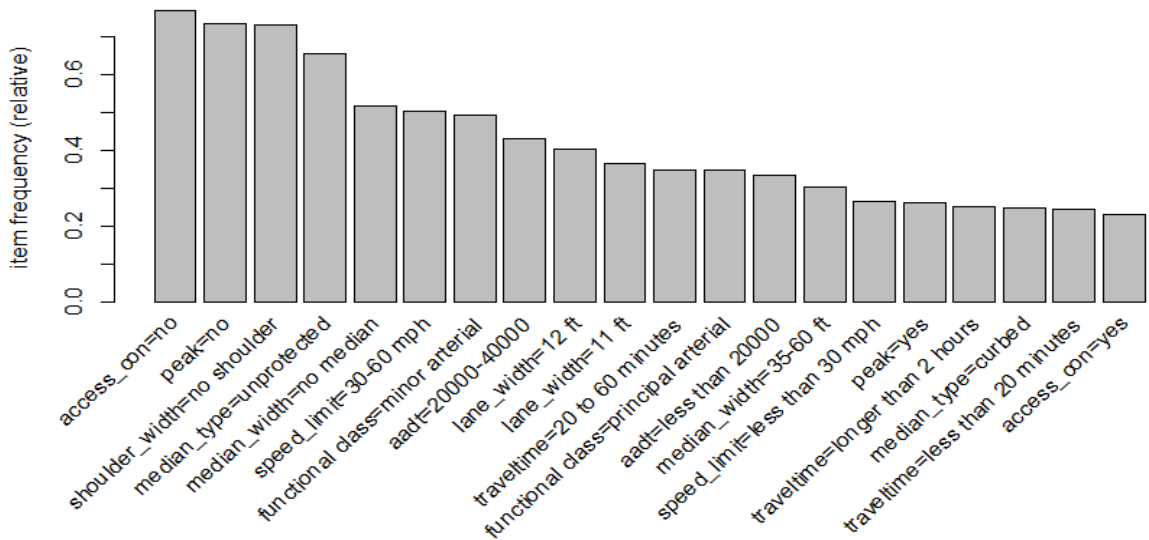

**Figure 2 Items Frequency Plot of Trivial Near-crash Events**

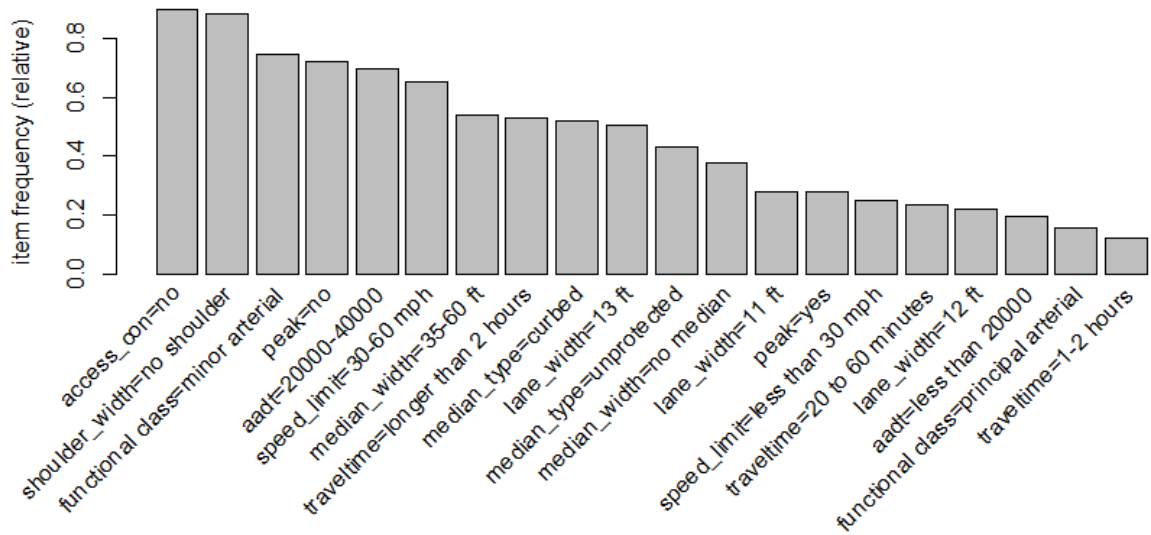

**Figure 3 Items Frequency Plot of Non-trivial Near-crash Events**

5.2. Model Performance and Descriptive Statistics of the Parameters

The mining setting is supervised data mining due to the major objective of this study is to mine meaningful association rules for trivial and non-trivial near-crash events. There are 475 rules for trivial near-crash events and 430 rules for non-trivial near-crash events. Table 4 describes the general information of three core parameters about the mined association rules. Their mean, min, and max value are reported.

**Table 4 Summary of the parameter values of mined rules**

| Consequent | # rules | Support | | | Confidence | | | Lift | | |
|---|---|---|---|---|---|---|---|---|---|---|
| | | Mean | Min. | Max. | Mean | Min. | Max. | Mean | Min. | Max. |
| **Trivial near-crash** | 475 | 0.14 | 0.10 | 0.40 | 0.53 | 0.36 | 0.81 | 0.91 | 0.62 | 1.40 |
| **Non-trivial near-crash** | 430 | 0.18 | 0.10 | 0.36 | 0.59 | 0.30 | 0.68 | 1.42 | 0.70 | 1.62 |

Figure 4 comprises two scatter plots, two-key plots of the parameters of all mined association rules. Order 3, order 4, and order 5 are marked in different colors to represent rules with various numbers of items. The purpose of the scatter plot is to provide an overview of the model performance. The distributions of the support and confidence value in the two plots indicate that a majority of rules have relatively high values of support and confidence.

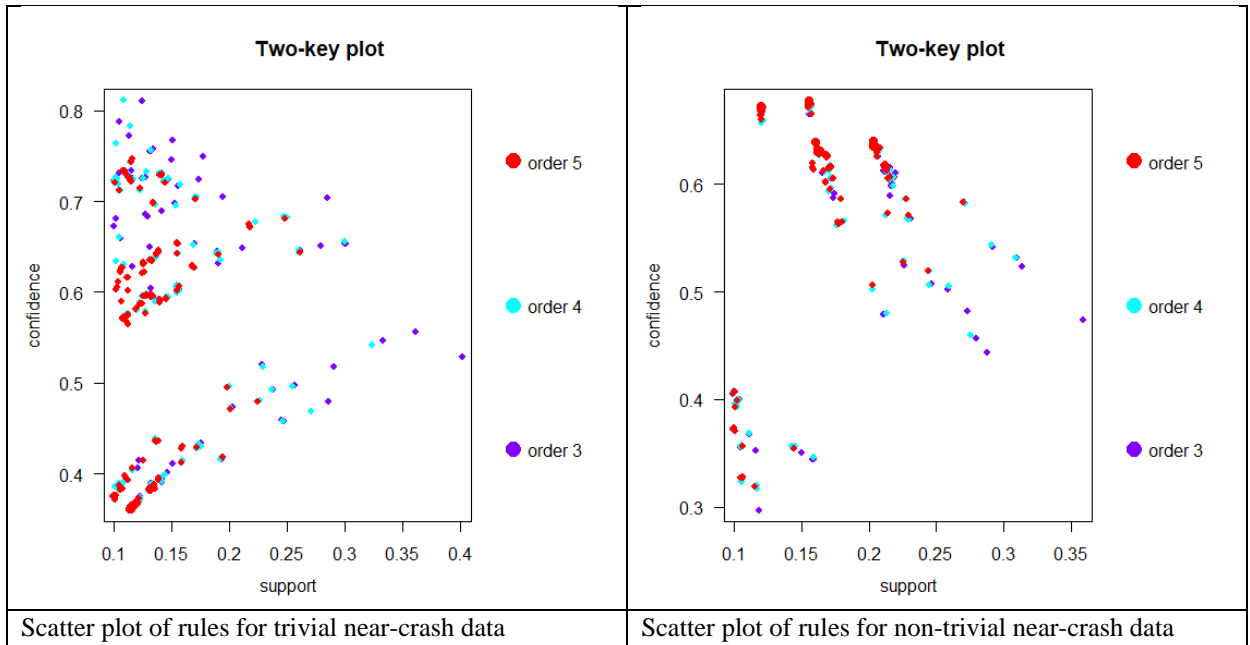

| Scatter plot of rules for trivial near-crash data | Scatter plot of rules for non-trivial near-crash data |

**Figure 4 Scatter plots of the parameters of mined association rules**

5.3. Understand Patterns from Association Rules of Trivial and Non-trivial Near-crash Events

Twenty-five rules of each type of near-crash event are selected based on the ranking of the Lift value. These rules are presented in Table 5 and Table 6 for trivial near-crash and non-trivial near-crash, respectively. In each table, 3-item rules and 4-5-item rules are presented separately on the ranking of the Lift value.

*5.2.1. Patterns for Trivial Near-crash events*

Table 5 lists 25 selected rules for trivial near-crash events with the highest *Lift* values from 475 mined rules. The first 4-5-item rule for trivial near-crash event is *{functional class = principal arterial + median_type = unprotected + peak = no}*. This rule has the highest *Lift* value among the 475 association rules. The assertion is that roads classified as principal arterial without median separating the opposing traffic are where the trivial near-crash events frequently occurred, especially during the non-peak hours. The corresponding indices follow this rule are *Support* = 10.8%, *Confident* = 81.1% and *Lift* = 1.396. As introduced in the methodology section, the value of *Support* indicates 10.8% of trivial near-crash events having these three components, i.e., driving on principal arterial without protected median during non-peak hours. The value of *Confidence* denotes that out of all near-crash events containing the three items, 81.1% of them are trivial near-crash events. In this dataset, the percentage of trivial near-crash events having these three items was 1.396 times higher than the percentage of trivial near-crash events in the overall dataset.

**Table 5 Top 25 Association Rules for Trivial Near-crash Events**

| Rules | Antecedent | | | | S | C | L | CT |
|---|---|---|---|---|---|---|---|---|
| | Item 1 | Item 2 | Item 3 | Item 4 | | | | |
| 3-item rules (Consequent item – Trivial Near-crash) | | | | | | | | |
| 1 | functional class = principal arterial | peak = no | -- | -- | 0.125 | 0.811 | 1.396 | 120 |

| # | Item 1 | Item 2 | Item 3 | Item 4 | S | C | L | CT |
|---|--------|--------|--------|--------|---|---|---|----|
| 2 | peak = no | traveltime = less than 20 minutes | -- | -- | 0.104 | 0.787 | 1.355 | 100 |
| 3 | median_type = unprotected | traveltime = less than 20 minutes | -- | -- | 0.113 | 0.771 | 1.328 | 108 |
| 4 | lane_width = 12 ft | peak = no | -- | -- | 0.153 | 0.768 | 1.323 | 146 |
| 5 | access_con = yes | lane_width = 12 ft | -- | -- | 0.134 | 0.757 | 1.304 | 128 |
| 6 | access_con = yes | speed_limit = larger than 60 mph | -- | -- | 0.133 | 0.756 | 1.301 | 127 |
| 7 | functional class = principal arterial | median_type = unprotected | -- | -- | 0.177 | 0.748 | 1.287 | 169 |
| 8 | functional class = principal arterial | lane_width = 12 ft | -- | -- | 0.149 | 0.745 | 1.282 | 143 |
| 9 | access_con = no | traveltime = less than 20 minutes | -- | -- | 0.115 | 0.733 | 1.262 | 110 |
| 10 | shoulder_width = no shoulder | traveltime = less than 20 minutes | -- | -- | 0.104 | 0.73 | 1.256 | 100 |
| 11 | speed_limit = 30-60 mph | aadt = less than 20,000 | -- | -- | 0.113 | 0.73 | 1.256 | 108 |
| 12 | functional class = principal arterial | shoulder_width = no shoulder | -- | -- | 0.124 | 0.726 | 1.249 | 119 |
| 13 | median_type = unprotected | speed_limit = 30-60 mph | -- | -- | 0.146 | 0.725 | 1.249 | 140 |
| 14 | lane_width = 11 ft | aadt=less than 20,000 | -- | -- | 0.126 | 0.725 | 1.247 | 121 |
| 15 | functional class = principal arterial | median_width = no median | -- | -- | 0.101 | 0.724 | 1.246 | 97 |
| **4-5-item rules (Consequent item – Trivial Near-crash)** | | | | | | | | |
| 16 | functional class = principal arterial | median_type = unprotected | peak = no | -- | 0.108 | 0.811 | 1.396 | 103 |
| 17 | lane_width = 12 ft | median_type = unprotected | peak=no | -- | 0.114 | 0.784 | 1.35 | 109 |
| 18 | access_con = no | median_type = unprotected | traveltime = less than 20 minutes | -- | 0.101 | 0.764 | 1.315 | 97 |
| 19 | median_type = unprotected | speed_limit = 30-60 mph | peak = no | -- | 0.117 | 0.747 | 1.285 | 112 |
| 20 | median_width = no median | speed_limit = 30-60 mph | peak = no | -- | 0.115 | 0.743 | 1.279 | 110 |
| 21 | functional class = principal arterial | lane_width = 12 ft | median_type = unprotected | -- | 0.13 | 0.734 | 1.263 | 124 |
| 22 | lane_width = 11 ft | median_width = no median | aadt = less than 20,000 | -- | 0.109 | 0.732 | 1.261 | 104 |
| 23 | access_con = no | shoulder_width = no shoulder | traveltime = less than 20 minutes | -- | 0.102 | 0.726 | 1.249 | 98 |
| 24 | functional class = principal arterial | shoulder_width = no shoulder | median_type = unprotected | -- | 0.116 | 0.725 | 1.249 | 111 |
| 25 | shoulder_width = no shoulder | median_width = no median | aadt = less than 20,000 | peak = no | 0.112 | 0.728 | 1.253 | 107 |

Note* S-support, C-confidence, L-lift, CT-count

To summarize the rules for trivial near-crash events, the rules indicate that roadways classified as principal arterial (rule1, 7, 8, 15, 16, 21, 24) with the absence of median (rule 1, 7, 11, 10, 13, 16, 17, 18, 19, 20, 22, 25) or shoulder (rule 10, 12, 23- 25) are highly associated with trivial near-crash events. Moreover, trivial near-crash events always occur in a relatively short trip less than 20 minutes (rule 2, 3, 9, 10, 18, 23) during the non-peak hour. It is interesting to discover that these trivial near-crash events predominantly happen during these relatively short trips on the principal arterials, which have a moderate speed limit between 30 to 60 mph (rule 11, 13, 19, 20) and do not have the protection of rigid median to separate the opposing traffic. One explanation could be drivers are less alert while driving relatively short trips. With the complexity of traffic conditions on principal arterials, trivial near-crash events are more likely to occur due to, e.g., sudden and unexpected stops caused by congestions or signalized intersections.

Meanwhile, this complexity on the principal arterials could be the reason that these near-crash events did not become non-trivial near-crash events or even crash events due to the expectations of these sudden stops or lane change behaviors. Moreover, items like the absence of a median or shoulders are found in 13 out of 25 rules, and the majority of them co-exist with the item "functional class = principle arterial." This observation further supported the hypothesis that the combination of items like principle arterial, absence of a median or shoulders, relatively short trips during non-peak hours triggers the possibility of the occurrence of trivial near-crash events. However, this combination also prevents the trivial near-crash events from becoming non-trivial crash events or even crashes because these unfavorable driving environments tend to alert drivers to maintain a relatively high level of attention to the driving tasks and to drive a relatively short trip to avoid more severe consequences.

### 5.2.2. Patterns for Non-Trivial Near-crash events

Table 6 lists 25 top rules for non-trivial near-crash events with the highest *Lift* values from 430 mined rules. The first 4-5-item rule in the table is *{lane_width = 13 ft + median_type = curbed + traveltime = longer than 2 hours}* with the following corresponding indices: *Support* = 15.6%, *Confidence* = 67.7% and *Lift* = 1.616. This rule asserts that the combination of a wide lane width of 13 feet, the presence of a curbed median, and relatively long trips (longer than 2 hours) are highly associated with non-trivial near-crash events. The *Support* value indicates that 15.6 percent of non-trivial near-crash events containing these three items. The value of *Confidence* shows that in the whole dataset, for the events containing these three items, 67.7% of them are non-trivial near-crash events. The *Lift* value also suggests the percentage of all non-trivial near-crash events having these three items was 1.616 times over the percentage of the non-trivial near-crash events in the overall dataset.

#### Table 6 Top 25 Association Rules for Non-trivial Near-crash Events

| Rules | Antecedent | | | | S | C | L | CT |
|---|---|---|---|---|---|---|---|---|
| | Item 1 | Item 2 | Item 3 | Item 4 | | | | |
| 3-item rules (Consequent item – Non-trivial Near-crash) | | | | | | | | |
| 1 | median_type = curbed | traveltime = longer than 2 hours | -- | -- | 0.156 | 0.674 | 1.609 | 149 |
| 2 | lane_width = 13 ft | traveltime = longer than 2 hours | -- | -- | 0.158 | 0.674 | 1.609 | 151 |
| 3 | median_width = 35-60 ft | traveltime = longer than 2 hours | -- | -- | 0.157 | 0.667 | 1.591 | 150 |
| 4 | median_width = 35-60 ft | aadt = 20,000-40,000 | -- | -- | 0.204 | 0.637 | 1.521 | 195 |
| 5 | lane_width = 13 ft | median_type = curbed | -- | -- | 0.204 | 0.635 | 1.516 | 195 |
| 6 | lane_width = 13 ft | aadt = 20,000-40,000 | -- | -- | 0.205 | 0.634 | 1.514 | 196 |

| # | item 1 | item 2 | item 3 | item 4 | S | C | L | CT |
|---|---|---|---|---|---|---|---|---|
| 7 | median_type = curbed | aadt = 20,000-40,000 | -- | -- | 0.206 | 0.633 | 1.512 | 197 |
| 8 | functional class = minor arterial | lane_width = 13 ft | -- | -- | 0.207 | 0.631 | 1.505 | 198 |
| 4-5-item rules(Consequent item – Non-trivial Near-crash) | | | | | | | | |
| 9 | lane_width = 13 ft | median_type = curbed | traveltime = longer than 2 hours | -- | 0.156 | 0.677 | 1.616 | 149 |
| 10 | shoulder_width = no shoulder | median_width = 35-60 ft | traveltime = longer than 2 hours | -- | 0.156 | 0.671 | 1.602 | 149 |
| 11 | lane_width = 13 ft | peak = no | traveltime = longer than 2 hours | -- | 0.121 | 0.671 | 1.6 | 116 |
| 12 | median_type = curbed | peak = no | traveltime = longer than 2 hours | -- | 0.12 | 0.669 | 1.596 | 115 |
| 13 | median_width = 35-60 ft | peak = no | traveltime = longer than 2 hours | -- | 0.121 | 0.659 | 1.573 | 116 |
| 14 | lane_width = 13 ft | median_type = curbed | aadt = 20,000-40,000 | -- | 0.204 | 0.639 | 1.526 | 195 |
| 15 | lane_width = 13 ft | median_type = curbed | peak = no | -- | 0.161 | 0.639 | 1.525 | 154 |
| 16 | median_width = 35-60 ft | peak = no | aadt = 20,000-40,000 | -- | 0.161 | 0.636 | 1.519 | 154 |
| 17 | functional class = minor arterial | speed_limit = 30-60 mph | aadt = 20,000-40,000 | -- | 0.208 | 0.634 | 1.512 | 199 |
| 18 | lane_width = 13 ft | peak = no | aadt = 20,000-40,000 | -- | 0.162 | 0.633 | 1.51 | 155 |
| 19 | functional class = minor arterial | shoulder_width = no shoulder | lane_width = 13 ft | -- | 0.207 | 0.633 | 1.51 | 198 |
| 20 | functional class = minor arterial | aadt = 20,000-40,000 | traveltime = longer than 2 hours | --- | 0.206 | 0.631 | 1.507 | 197 |
| 21 | median_type = curbed | peak = no | aadt = 20,000-40,000 | -- | 0.162 | 0.63 | 1.504 | 155 |
| 22 | lane_width = 13 ft | median_type = curbed | peak = no | traveltime = longer than 2 hours | 0.12 | 0.673 | 1.605 | 115 |
| 23 | shoulder_width = no shoulder | median_width = 35-60 ft | peak = no | traveltime = longer than 2 hours | 0.12 | 0.665 | 1.586 | 115 |
| 24 | functional class = minor arterial | speed_limit = 30-60 mph | peak = no | aadt = 20,000-40,000 | 0.164 | 0.633 | 1.511 | 157 |
| 25 | shoulder_width = no shoulder | lane_width = 13 ft | speed_limit = 30-60 mph | peak = no | 0.164 | 0.631 | 1.505 | 157 |

*Note* S-support, C-confidence, L-lift, CT-count

In these top 25 association rules for non-trivial near-crash events, the dominant items are *median_width = 35-60 ft (15 out of 25 rules), lane_width = 13 ft (12 out of 25 rules), travel time = longer than 2 hours (11 out of 25 rules), aadt = 20,000-40,000 vpd (10 out of 25 rules), and median_type= curbed (9 out of 25 rules).* The existence of a median, wide lane width, relatively high traffic volume, and relatively long trips are critical factors highly associated with non-trivial near-crash events. A combination of two or three items out of these four dominant items was found in 21 out of 25 rules with the highest *Lift* values. For example, the presence of median and travel time longer than 2 hours are two items showing in 8 out of the top 10 rules (rule 1, 2, 4, 5, 7, 8, 9, 10) with the highest *Lift* values in Table 6. The mined rules suggest the combination of longer trips and the presence of a median contributes even more to non-trivial near-crash incidents. Long trips often cause fatigue and impatience that may lead to severe driving incidents. The presence of median curbs or wide medians provides a barrier or buffer between the driver and opposing traffic. Such driving environments may cause drivers to be less cautious while driving. Fatigue from a long drive coupled with careless driving behaviors may have led to the surge in non-trivial near-crash events. Frrom Table 6, these roadways contain features such as 35 – 60 feet median width, 13 feet lane width, and 20,000-40,000 vpd, which often are associated with a relatively higher functional class such as interstate highways. In other words, the non-trivial crash events are more likely to occur on the roadways with higher functional classes.

Additionally, several rules (rule 17, 19, 20, 24) denote that minor arterials could also be a place where non-trivial near-crash events often occurred. In these four rules, the combination of functional class = minor arterial and aadt = 20,000-40,000 vpd shows in every rule. This indicates that roadways with a relatively lower functional class could also be the place where non-trivial near-crash events often occur if the traffic volume is relatively heavy. Based on the traffic volume criteria for various functional class roadways provided by Federal Highway Administration (FHWA), the expected traffic volume for rural minor arterial is 1,500-6,000 vpd and for urban minor arterial is 3,000-14,000 vpd (Administration, 2013). The traffic volume of 20,000-40,000 vpd for the minor arterial can be considered a very heavy traffic volume. This explains heavier daily traffic volume could trigger more non-trivial near-crash events.

### 5.2.3. Comparison of the Patterns of Trivial and Non-Trivial Near-Crash Events

The authors also compared the patterns in the two sets of rules. Trivial near-crash events are often to be associated with the occurrence on roadways with relatively lower functional class, without a median or shoulders, whereas non-trivial near-crash events are mostly to be associated with the occurrence on roadways with relatively higher functional class with the presence of median. Another finding is the trivial near-crash events often associated with relatively short trips (less than 20 minutes in this dataset). On the contrary, the non-trivial near-crash events are more likely to be associated with trips longer than 2 hours. The results also indicate the non-trivial near-crash events are also associated with occurring on the roadways with a relatively lower functional class if the roadway has a relatively higher daily traffic volume.

## 6.  Findings and Discussions

To summarize this study, the key findings are listed as follows:

- In general, near-crash events are more likely to be associated with road segments without access control, absence of shoulders, unprotected median, driving during non-peak hours, 30-60 mph posted speed limit and trivial near-crash event. The first four findings are consistent with the fact that access control, shoulder provision, protected median, and lower volume levels are often associated with improved roadway safety to reduce the odds of safety-critical events (AASHTO, 2010). Roadways with a 30-60 mph speed limit in the study are often principal arterials, which often have less ideal driving conditions regarding the first three aspects (Sawalha and Sayed, 2001). Because these complex driving conditions are coupled with relatively lower driving speeds, the associated near-crash events tend to be trivial.
- By comparing the patterns associate with the trivial and non-trivial near-crash events, the results found that trivial near-crashes often occurs on relatively lower functional class roadways without a median or shoulders, whereas the non-trivial near-crashes often occur on roadways with a relatively higher functional class with the presence of median. These findings revealed that severer near-crash events occur on the roadway with fewer interruptions and more protections, such as the median. Higher operational speed on these roadways (Ivan et al., 2009) could be one reason. And less attention or alert from drivers due to the more secure driving environment could be another reason behind this pattern. These findings are also consistent with that average speed prior to a crash is highly correlated with severity outcomes (Arvin et al., 2019).

- The results also indicate that the non-trivial near-crashes are more likely to occur on the longer trips. Tiredness and boredness could be a trigger of irrational driving behavior (Cummings et al., 2001).

- The analysis also revealed that the non-trivial near-crash also could associates with a relatively lower function class roadway with a relatively higher daily traffic volume. Higher traffic volume could introduce more opportunities for near-crash events. Moreover, higher traffic volume often associates with congestions. The corridor with congestions issues increases the uncertainty of the driving environment. The slow-moving traffic and unexpected braking behavior from the leading vehicles often trigger severe near-crash events. This is consistent with the fact that higher volume levels are associated with a higher driving risk (AASHTO, 2010).

## 7. Conclusions

This study explored the associations between the trivial and non-trivial longitudinal near-crash events and road geometric/trip features in a naturalistic driving dataset by using association rules mining algorithm. To thoroughly understand the common and unique patterns of the trivial and non-trivial events, this study mined association rules for the trivial and non-trivial near-crash events separately and generated much clearer insights by comparing the patterns from the two sets of rules. The current study has two major contributions. First, it used a unique dataset that can be used to understand the near-crash event patterns from a naturalistic driving dataset. Second, this study developed a framework of the application of association rules mining to identify the unknown and interesting rules. The mined pattern not only just associates one category with the occurrence of the near-crash event but also discovers the associates among multiple categories. The traditional statistical methods could calculate the marginal effects of certain variables on the near-crash or crash events. The results provided from association rules explore the relationship from a completely different angle. One category, such as a low functional class roadway, may not directly be associated with the occurrence of near-crash events, but it could become a dominant factor triggering the near-crash event if combining with other factors, such as a high traffic volume, frequently congested, with a median and a wide shoulder. Association rule mining is a powerful algorithm but not widely used in the transportation field. The framework can be used in other similar surrogate approaches.

The trivial near-crash events are more likely to be associated with the occurrence on roadways with a relatively lower functional class and the absence of a median or shoulders. Trivial near-crash events are also associated with relatively short trips. The combinations of these geometry and trip features are dominantly presented in the mined association rules for trivial near-crash events. On the contrary, the non-trivial near-crash events have a higher probability of being associated with the occurrence on roadways with a relatively higher functional class and with the presence of a median. The non-trivial also occur often on the roadways with a relatively lower functional class but with a relatively heavier daily traffic volume.

Longer trips are also highly associated with non-trivial near-crash events. The combinations of these features for non-trivial events are dominantly presented in the mined association rule sets.

Identifying factors contributing to the near-crash events is a proven surrogate approach to study the crash-related transportation safety issues. The findings of this study could help researchers and engineers to understand near-crash events and identify the near-crash event hot spots and further provide countermeasures to mitigate the safety risks. For example, the findings indicate the combination of a curbed median and a 13 feet wide lane triggers more non-trivial near-crash events, especially for roadways with a relatively lower functional class than freeways. To reduce the risks of crashes, the transportation agencies, law enforcement, and research institutes may investigate more into these areas and modify these features to mitigate the crash risks. The results also indicate roadways with a relatively lower functional class but having a heavy daily traffic volume tend to have a higher rate of non-trivial near-crash events. The association suggested by this finding could be an incentive for local transportation safety agencies to focus more on roadways with a relatively lower functional class and high rates of crash or near-crash occurrences.

There are three limitations in this study that are worth mentioning. Firstly, the association rule mining process relies heavily on the parameters *Support*, *Confidence*, and *Lift*, and different settings of the parameters could yield a number of different rules. The authors went through a large number of iterations during the parameter tuning process to locate the proper parameters to secure the stable performance of the model. Future studies could improve the process of optimizing these parameters. There are several other performance measures (e.g., lift increases, rules power factor, leverage) that could be applied to generate rules. For example, the lift increase (LIC) could use one item's lift value as a start point, and then rules with more items could be selected if their lift values pass a certain threshold (López et al., 2014; Montella et al., 2020, 2012, 2011). Future studies can be improved by adopting other available performance measurements and select rules based on these alternative measurements. Secondly, one concern lies in the HPMS dataset, where a few important roadway features (e.g., type of shoulders) have many missing values. These features were excluded from the analysis due to the missing values, but they might have played a role in the near-crash event occurrence. One possible solution to this is to use the imputation method, which was not applied in this study. The study could also be improved if a more comprehensive road inventory dataset becomes available. Thirdly, this study only considered near-crash events that involved another vehicle. The amount of data for near-crash events involving objects is lacking. In future studies, this type of near-crash event could be included if data becomes available.

**Acknowledgment**